\documentclass{article}
\usepackage{spconf,amsmath,graphicx}

\title{Co-attentional transformers for story-based video understanding}
\name{Bj\"orn Bebensee$^1$, Byoung-Tak Zhang$^2$\thanks{This work was partly supported by the Institute for Information \& Communications Technology Promotion (2015-0-00310-SW.StarLab, 2017-0-01772-VTT, 2018-0-00622-RMI, 2019-0-01367-BabyMind) and Korea Institute for Advancement Technology (P0006720-GENKO) grant funded by the Korean government.}}
\address{$^1$Deptartment of Computer Science and Engineering \quad $^2$Artificial Intelligence Institute (AIIS)\\
Seoul National University, Seoul, Republic of Korea\\
\{bebensee, btzhang\}@snu.ac.kr }
\usepackage{subfig}
\usepackage{booktabs} 
\usepackage{amsmath}
\usepackage{caption}
\usepackage{amssymb}
\usepackage{placeins}

\begin{document}
%
\maketitle
\begin{abstract}
Inspired by recent trends in vision and language learning, we explore applications of attention mechanisms for visio-lingual fusion within an application to story-based video understanding. Like other video-based QA tasks, video story understanding requires agents to grasp complex temporal dependencies. However, as it focuses on the narrative aspect of video it also requires understanding of the interactions between different characters, as well as their actions and their motivations. We propose a novel co-attentional transformer model to better capture long-term dependencies seen in visual stories such as dramas and measure its performance on the video question answering task. We evaluate our approach on the recently introduced DramaQA dataset~\cite{choi2020dramaqa} which features character-centered video story understanding questions. Our model outperforms the baseline model by 8 percentage points overall, at least 4.95 and up to 12.8 percentage points on all difficulty levels and manages to beat the winner of the DramaQA challenge.
\end{abstract}
\section{Introduction}
\label{sec:introduction}
Both computer vision and natural language processing have seen several breakthroughs in recent years leading to large progress in tasks combining both these modalities such as image retrieval, image captioning and visual question answering. In particular, Visio-lingual representation learning has made great progress benefiting these downstream tasks~\cite{lu2019vilbert,huang2020pixel,su2020vlbert,li2020unicoder,chen2019uniter}. As an extension of visual question answering in the temporal domain, video question answering adds this additional dimension to an already challenging problem and has thus received considerably less attention. Not only should models learn to utilize contextual information and references between the vision and language but also to perform multi-step and long-term reasoning in the temporal axis.

In video question answering an agent is presented with a video clip of a scene and has to infer the correct answer to a given question in natural language. Such a scene can consist either of a single shot or multiple shots from different angles or in different locations. While questions can be relatively simple, e.g. ``What is the woman holding?'', or more complex and require deeper understanding and multiple steps of reasoning, e.g. ``Why is the man in the overalls angry at the cyclist?''. Due to these temporal dependencies that need to be resolved and understood in order to answer more complex questions about the scene correctly, video question answering has remained a very challenging problem.

One such dataset for video question answering is the TVQA dataset which is build around short 60 to 90 seconds long video clips and questions bridging vision and language clues~\cite{lei2018tvqa}. Agents have to infer the answers by using multiple modalities (video frames, subtitle-based dialogue) as well as temporally localize the relevant part. TVQA+ adds additional bounding boxes and objects annotations that link them directly to visual concepts mentioned in questions and answers~\cite{lei2020tvqaplus}. While the TVQA(+) dataset seems to be a popular choice for evaluation of video question answering and video understanding models~\cite{yang2020bert,kim2019gaining,geng2020character}, it does not require story-level understanding. Most questions in the dataset only require the agent to attend to a short part of the video clip (15 seconds or less) due to its particular focus on temporal localization.

In this work we instead focus on story-based video understanding for long-term dependencies and a deeper understanding of characters' actions and intentions. In order to fuse vision and language in a meaningful way, we adopt a co-attentional transformer inspired by recent work in visual dialog~\cite{nguyen2019efficient}. We evaluate our approach on the recent DramaQA dataset which features questions closely centered around the narrative and characters of a TV drama along with character-level annotations. Unlike in TVQA, questions focus on longer-range character interactions and aim to capture story understanding on a deeper level at both the shot and the scene level.

We will first review work most closely related to our approach, introduce the evaluation dataset, and finally our model architecture and experimental results.


\begin{figure*}
    \centering
    \includegraphics[width=\textwidth]{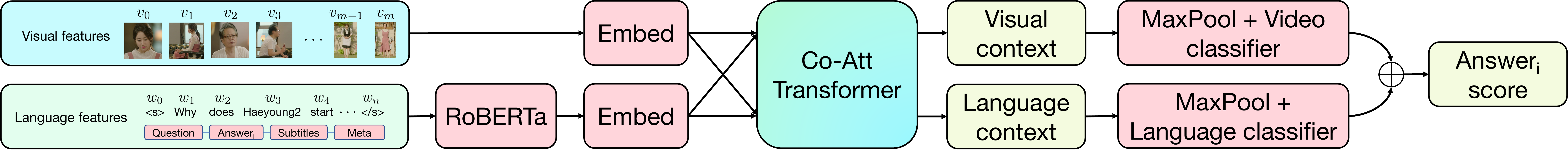}
    \caption{Our two-stream transformer model manages to learn cross-modal interactions through co-attention. The final answer score takes into account both the visual context attended to the language representations as well as the language representations attended to the visual bounding boxes.}
    \label{fig:architecture}
\end{figure*}

\section{Related Work}
\label{sec:relatedwork}
\noindent\textbf{Vision and language fusion.} Most work in vision and language fusion as well as visio-lingual representation learning focuses on single image-text pairs. Recently there has been a lot of work on learning better joint representations of vision and language especially, with many of these models reaching top performance in downstream tasks such as visual question answering. VL-BERT~\cite{su2020vlbert}, Unicoder-VL~\cite{li2020unicoder}, and UNITER~\cite{chen2019uniter} combine vision and language representations into a single stream that is input into a Transformer encoder. ViLBERT~\cite{lu2019vilbert} and LXMERT~\cite{tan2019lxmert} instead use separate streams for vision and language that are first separately encoded and then cross-encoded. Although we do not employ any form of pretraining tasks for representation learning int his work, this latter type of two-stream co-attention between modalities is most similar to our approach.

\noindent\textbf{Video representation learning.} Video representation learning for downstream tasks such as video captioning is a new but active area of research as well. Much like VL-BERT, VideoBERT~\cite{sun2019videobert} learns vision and language joint representations using a single stream of data with a joint encoder. VideoBERT is focused on instructional videos collected from YouTube for captioning and activity recognition. UniVL~\cite{luo2020univilm} similarly uses instructional videos for its pretraining task but instead employs a cross-encoder architecture.

\noindent\textbf{Video question answering.} In order to deal with additional challenges in video QA, Fan et al.~\cite{fan2019heterogeneous} introduce LSTM-based memory modules for video and question utilities to capture contextual information from the reference video. Inspired by the success of Transformers in NLP, recent work by Li et al.~\cite{li2019beyond} on spatiotemporal reasoning in short video clips introduces a positional self-attention block as an encoder for each modality stream along with a co-attention block. Instead of using video input directly, Yang et al.~\cite{yang2020bert} use a two-stream BERT model for video QA which encodes the semantic content of a video scene as the visual concept labels of the detected objects given by Faster R-CNN. 

\section{Problem formulation}
\label{sec:problem}
We formulate the problem in a similar manner as Choi et al. \cite{choi2020dramaqa}. That is, for a given question sequence $Q$ and given the reference video clip's script sequence $S$ along with a visual feature stream $V$ and visual metadata $M$, we want to infer the correct answer sequence $A_i \in \{A_1, \ldots, A_5 \}$. The visual feature stream consists of a series of region features extracted from the characters' full body bounding box for each frame in the reference video.

\subsection{Dataset}
To conduct our experiments we choose the DramaQA dataset which aims to benchmark video story understanding and focuses in particular on story-level questions (rather than shorter dependencies seen in the TVQA dataset)~\cite{choi2020dramaqa}. In particular, it is based on the Korean TV show ``Another Miss Oh'', consisting of 23\,928 video clips, of which 803 are scene-level and 23\,125 are shot-level clips, and spanning 18 episodes in total. The questions are multiple-choice with five possible answers to choose from and can be categorized into four difficulty levels. Depending on the level of difficulty higher-level understanding of the scene may be necessary to answer correctly. The difficulty ranges from level 1 which only requires a single supporting fact to answer the question (i.e. ``Who is holding the phone?''), level 2 which requires multiple such supporting facts to level 3 and 4 which require to reason across the temporal axis and to understand causal relationships between multiple supporting facts. Along with the image frames the dataset provides the coreference-resolved dialogue scripts for each video clip. Additionally, it provides bounding boxes for characters appearing in each frames with visual metadata annotations containing name, behavior, emotion i.e. ``$\{\text{Doegi},\text{standing up},\text{surprise}\}$''.

\section{Proposed method}
\label{sec:method}

\begin{figure}
    \centering
    \includegraphics[width=.5\linewidth]{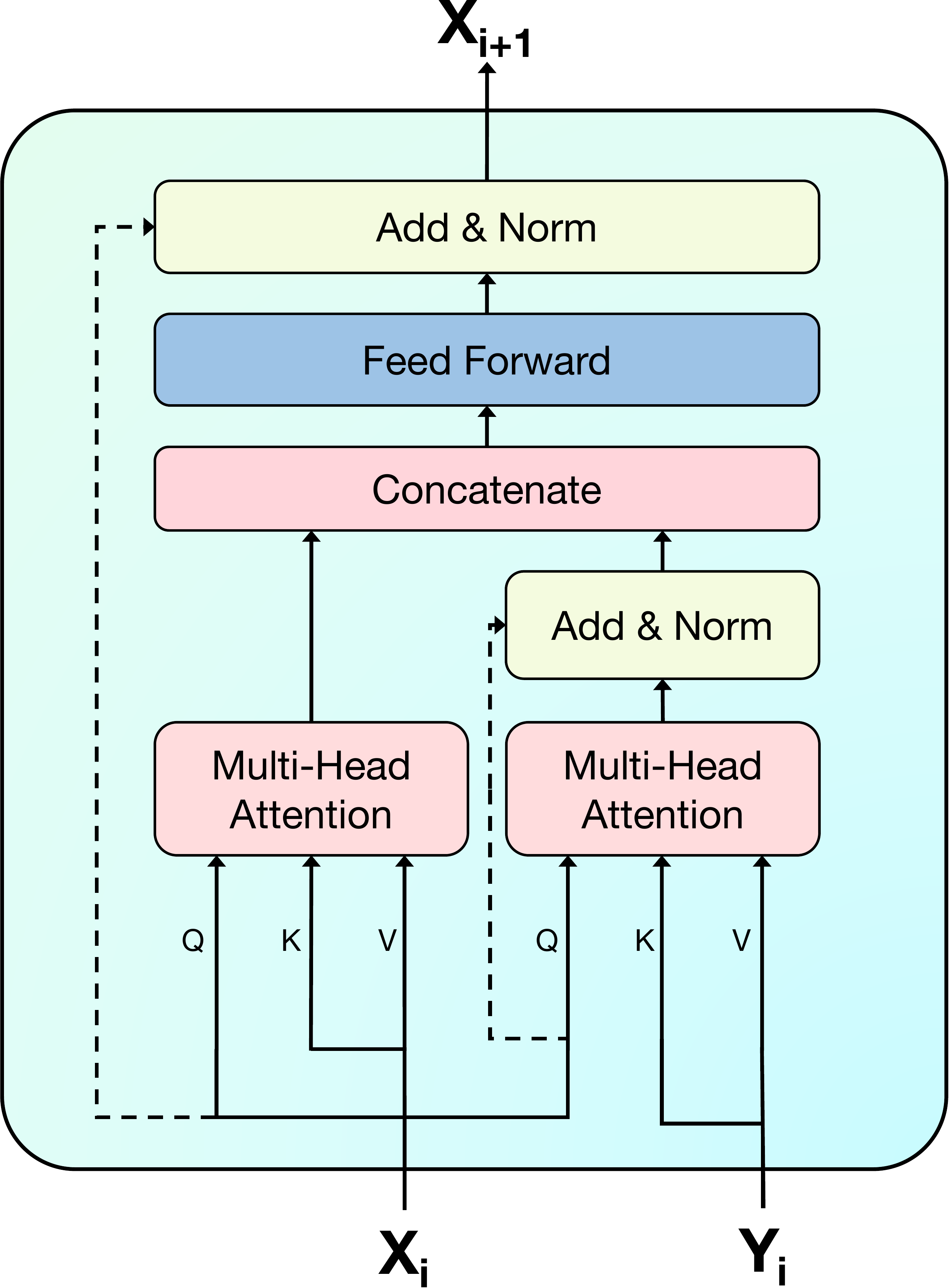}
    \caption{A single co-attention block attends a target utility $X$ to a source utilities Y. We use two of these blocks per layer, attending both the visual representations to the language representations and vice versa. Given input representations $L^{(k-1)}_{i}, V^{(k-1)}_{i}$ for answer option $i$, we obtains new utilities $L^{(k)}_{i}, V^{(k)}_{i}$ that are attended to one another.}
    \label{fig:coatt}
\end{figure}

Our method takes inspiration from recent advances in vision and language fusion especially with applications to visual question answering and visual dialog. Recent work by Nguyen et al. \cite{nguyen2019efficient} on the visual dialog task introduced a new type of co-attention layer for three or more input modalities (image, question, dialog history) with fewer trainable weights, that can be stacked in order to better integrate dependencies between many different utilities. While ViLBERT's co-attention layer only takes two modalities as input, it co-attends them in a similar fashion but uses more powerful language representations from BERT. Video question answering requires agents to fuse vision and language features to infer the correct answer to a question in the same fashion and can benefit from a similar co-attention layer as well.

\subsection{Model architecture}

Following Choi et al.~\cite{choi2020dramaqa} we use question features, subtitle features, answer features, meta features (behavior and emotion) as well as visual features. In order for the language model to infer ``who is doing what and feeling how'' from the meta features more easily we transform them to sentences, i.e. ``Doegi is standing up and feeling sadness''. To encode these language token sequences to obtain language representations we use a pretrained RoBERTa model~\cite{liu2019roberta}, a variant of the widely successful BERT model that achieves significantly better performance with the same architecture but tuned hyperparameters and more training data. We leverage the powerful RoBERTa model to learn better visio-lingual representations for video story understanding. 
First, we fuse all textual inputs as follows; for the $i$-th answer option we obtain the concatenation

\begin{equation}
l_i = [Q; S; M; A_i].
\end{equation}

During concatenation, we separate the tokenized sequences using RoBERTa's start and end of sentence tokens as follows:

\begin{equation}
\langle s  \rangle \ Q \ \langle /s \rangle \quad \langle s \rangle \ S \ \langle /s \rangle \quad \langle s \rangle \ M \ \langle /s \rangle \quad \langle s \rangle \ A_i \ \langle /s \rangle
\end{equation}

Although RoBERTa has not been pretrained for more than two sentence types, this separation will aid the model in differentiating between the different input types. Fusing all textual inputs gives us a total of 5 language token sequences $l = \{l_1, \ldots, l_5 \}$ representing the respective question-answer pairs along with the respective subtitles and meta features.  Given the token sequence $l_i$ of length $n$, we can now use RoBERTa to obtain a language representation $\mathbf{L_i} \in \mathbb{R}^{n \times d_L}$ where $d_L$ is the hidden size of the text representation.

\begin{equation}
\mathbf{L_i} = \text{RoBERTa}(l_i)
\end{equation}

Finally, we use a linear layer projection to obtain representations in a joint visio-lingual embedding space of dimension $d$.

To obtain visual representations, we extract image region features of the main characters for each frame from the annotated bounding boxes using a pretrained image feature extractor such as ResNet-152. We also use a linear layer to project these dimensions from the image feature dimension $d_V$ to the joint visio-lingual embedding space of dimension $d$.

In order to capture both dependencies between the two modalities and within the temporal axis, we use a co-attentional transformer (see Figure \ref{fig:coatt}) from to obtain vision and language context representations (for each answer option) that have been fused with the other modality. Given a tuple of vision and language representations in the joint embedding space $(L^{(0)}_{i}, V^{(0)}_{i}) \in \mathbb{R}^{n \times d} \times \mathbb{R}^{m \times d}$ we obtain the attended representations using two co-attention blocks as follows:

\begin{table}
    \centering
    \resizebox{\linewidth}{!}{%
    \begin{tabular}{@{}lcc@{}}
        \toprule 
        \quad                &  Overall & Difficulty Avg. \\ \toprule
        Dual Matching Multistream \cite{choi2020dramaqa} & 0.7201 & 0.6421  \\ \midrule
        Discriminative & 0.7207 & 0.6654 \\ \midrule\midrule
        Question + Answer & 0.5630 & 0.5566 \\ \midrule
        \quad + Subtitles + Video + Meta & \textbf{0.7711} & \textbf{0.7256} \\ \bottomrule
    \end{tabular}
    }
    \caption{Evaluation results on the DramaQA validation set. We compare our model to an additional discriminative baseline which adops the discriminative decoder from the visual dialog task on top of the DMM baseline~\cite{das2017visual} as well as a simple RoBERTa question + answer baseline.}
    \label{table:valset}
\end{table}

\addtolength{\tabcolsep}{3pt} 
\begin{table*}
    \centering
    \begin{tabular}{@{}lcccc|cc@{}}
        \toprule
        Model & Level 1 & Level 2 & Level 3 & Level 4 & \textbf{Overall} & \textbf{Difficulty Avg.} \\ \toprule
        Dot product (Q+A) \cite{choi2020dramaqa} & 0.5146 & 0.4766 & 0.4146 & 0.5038 & 0.4898 & 0.4774 \\ \midrule
        MLP (Q+A) \cite{choi2020dramaqa} & 0.3029 & 0.2773 & 0.2818 & 0.2743 & 0.2895 & 0.2841 \\ \midrule
        Dual Matching Multistream \cite{choi2020dramaqa}  & 0.7569 & 0.7110 & 0.5603 & 0.5592 & 0.6924 & 0.6469  \\ \midrule \midrule
        Ours & \textbf{0.8064} & \textbf{0.7843} & \textbf{0.6846} & \textbf{0.6870} & \textbf{0.7724} & \textbf{0.7406} \\ \bottomrule
    \end{tabular}
    \caption{Evaluation results on the DramaQA test set by question logic level, overall and average across the difficulties. Higher levels require more complex reasoning. }
    \label{table:testset}
\end{table*}
\addtolength{\tabcolsep}{-3pt} 

\begin{table*}
    \centering
    \begin{tabular}{@{}lllll|ll@{}}
        \toprule
        Team & Difficulty 1 & Difficulty 2 & Difficulty 3 & Difficulty 4 & \textbf{Overall} & \textbf{Difficulty Avg.} \\ \toprule
        GGANG & \textbf{0.81} & \textbf{0.79} & 0.64 & \textbf{0.70} & \textbf{0.77} & 0.73 \\ \midrule
        Sudoku & 0.78 & 0.74 & 0.68 & 0.67 & 0.75 & 0.72 \\ \midrule
        HARD KAERI  & 0.76 & 0.73 & 0.56 & 0.59 & 0.71 & 0.66 \\ \midrule \midrule
        Ours & \textbf{0.8064} & 0.7843 & \textbf{0.6846} & 0.6870 & \textbf{0.7724} & \textbf{0.7406} \\ \bottomrule
    \end{tabular}
    \caption{Comparison of our model's results with the winners of the DramaQA challenge 2020 held at ECCV 2020. Results are evaluated on the DramaQA test set. All results are rounded to two decimal places on the scoreboard. The winning critera was difficulty average.}
    \label{table:challenge}
\end{table*}

\begin{align}
    L^{(1)}_i &= \mathrm{CoAtt}(L^{(0)}_{i}, V^{(0)}_{i})\\
    V^{(1)}_i &= \mathrm{CoAtt}(V^{(0)}_{i}, L^{(0)}_{i})
\end{align}

Finally, we take the maximum along the sequence for both the visual and the langauge stream and use a linear classifier to obtain a vision and language answer score for each answer.

\begin{align}
    S_V &= \mathrm{max}(c_V)\\
    S_L &= \mathrm{max}(c_L)
\end{align}

The final scores are computed by simply taking the sum of the vision and language answer scores for each answer option:

\begin{equation}
    \text{Score} = S_V + S_L
\end{equation}

\section{Evaluation}
\label{sec:evaluation}

\subsection{Implementation details}
For our experiments we extract 2048-dimensional region features from the bounding box annotations using a ResNet-152 model that has been pretrained on ImageNet. For the linguistic stream we use $\text{RoBERTa}_\texttt{BASE}$ consisting of 12 transformer encoder layers with a hidden size of $d_L=768$ and 12 attention heads. We believe that the $\text{RoBERTa}_\texttt{LARGE}$ model may lead to better results but have chosen the $\texttt{BASE}$ model for our experiments due to resource constraints. We use a joint embedding space of dimension $d = 300$. We use a single co-attentional transformer layer with 6 attention heads. For the maximum length of the video input sequence we choose $M = 300$, for the maximum number of tokens in the text input sequence $N = 300$ and
We train the model using two Titan Xp GPUs with a batch size of 4 for a total of 5 epochs. We use the Adam optimizer with a learning rate of $10^{-4}$ and a weight decay of $10^{-5}$. To train the model we use a softmax and a cross-entropy loss on the predicted answer scores.

\subsection{Experimental results}

We compare our co-attentional transformer model against the baselines reported by Choi et al. \cite{choi2020dramaqa} on the test set. Results of our evaluation can be seen in Table \ref{table:testset}. Our model outperforms all baselines. We improve upon the Dual Matching Multistream baseline model on all difficulty levels; at least by about 5 percentage points and at most by 12.8 percentage points. Overall on the entire test set we can see a large improvement of 8 percentage points.

Across the difficulty levels we see the largest improvement over levels 3 and 4 where causal and long-term reasoning is necessary to infer the correct answer. We attribute this to the stronger fusion of vision and language resulting from the co-attentional transformer model, thus allowing the classifier to take advantage of more complex cross-modal clues. Additional results obtained on the validation data set can be seen in Table \ref{table:valset}. 

We also compare our model to the three winners of the DramaQA challenge at ECCV 2020. Although the overall evaluation results on the full dataset are very similar to the first place winner, our model outperforms the winning model on difficulty average, the winning criteria of the challenge.

\section{Conclusion}
\label{sec:conclusion}

We introduced a novel two-stream co-attention transformer architecture for story-based video understanding that successfully learns cross-modal relationships. We evaluated our architecture in a video question answering setting with character-centered annotations and questions  on the DramaQA dataset. Our model outperforms the Dual Matching Multistream baseline model on every difficulty level by at least 4.95 and up to 12.8 percentage points on higher difficulty levels that require more complex reasoning. 


\pagebreak

\bibliographystyle{IEEEbib}
\bibliography{refs}

\begin{thebibliography}{10}

\bibitem{choi2020dramaqa}
Seongho Choi, Kyoung-Woon On, Yu-Jung Heo, Ahjeong Seo, Youwon Jang, Seungchan
  Lee, Minsu Lee, and Byoung-Tak Zhang,
\newblock ``Drama{QA}: Character-centered video story understanding with
  hierarchical {QA},''
\newblock {\em arXiv preprint arXiv:2005.03356}, 2020.

\bibitem{lu2019vilbert}
Jiasen Lu, Dhruv Batra, Devi Parikh, and Stefan Lee,
\newblock ``Vilbert: Pretraining task-agnostic visiolinguistic representations
  for vision-and-language tasks,''
\newblock in {\em Advances in Neural Information Processing Systems}, 2019, pp.
  13--23.

\bibitem{huang2020pixel}
Zhicheng Huang, Zhaoyang Zeng, Bei Liu, Dongmei Fu, and Jianlong Fu,
\newblock ``Pixel-bert: Aligning image pixels with text by deep multi-modal
  transformers,''
\newblock {\em arXiv preprint arXiv:2004.00849}, 2020.

\bibitem{su2020vlbert}
Weijie Su, Xizhou Zhu, Yue Cao, Bin Li, Lewei Lu, Furu Wei, and Jifeng Dai,
\newblock ``Vl-bert: Pre-training of generic visual-linguistic
  representations,''
\newblock in {\em International Conference on Learning Representations}, 2020.

\bibitem{li2020unicoder}
Gen Li, Nan Duan, Yuejian Fang, Ming Gong, Daxin Jiang, and Ming Zhou,
\newblock ``Unicoder-vl: A universal encoder for vision and language by
  cross-modal pre-training.,''
\newblock in {\em AAAI}, 2020, pp. 11336--11344.

\bibitem{chen2019uniter}
Yen-Chun Chen, Linjie Li, Licheng Yu, Ahmed~El Kholy, Faisal Ahmed, Zhe Gan,
  Yu~Cheng, and Jingjing Liu,
\newblock ``Uniter: Learning universal image-text representations,''
\newblock {\em arXiv preprint arXiv:1909.11740}, 2019.

\bibitem{lei2018tvqa}
Jie Lei, Licheng Yu, Mohit Bansal, and Tamara Berg,
\newblock ``Tvqa: Localized, compositional video question answering,''
\newblock in {\em Proceedings of the 2018 Conference on Empirical Methods in
  Natural Language Processing}, 2018, pp. 1369--1379.

\bibitem{lei2020tvqaplus}
Jie Lei, Licheng Yu, Tamara Berg, and Mohit Bansal,
\newblock ``{TVQA}+: Spatio-temporal grounding for video question answering,''
\newblock in {\em Proceedings of the 58th Annual Meeting of the Association for
  Computational Linguistics}, Online, July 2020, pp. 8211--8225, Association
  for Computational Linguistics.

\bibitem{yang2020bert}
Zekun Yang, Noa Garcia, Chenhui Chu, Mayu Otani, Yuta Nakashima, and Haruo
  Takemura,
\newblock ``Bert representations for video question answering,''
\newblock in {\em The IEEE Winter Conference on Applications of Computer
  Vision}, 2020, pp. 1556--1565.

\bibitem{kim2019gaining}
Junyeong Kim, Minuk Ma, Kyungsu Kim, Sungjin Kim, and Chang~D Yoo,
\newblock ``Gaining extra supervision via multi-task learning for multi-modal
  video question answering,''
\newblock in {\em 2019 International Joint Conference on Neural Networks
  (IJCNN)}. IEEE, 2019, pp. 1--8.

\bibitem{geng2020character}
Shijie Geng, Ji~Zhang, Zuohui Fu, Peng Gao, Hang Zhang, and Gerard de~Melo,
\newblock ``Character matters: Video story understanding with character-aware
  relations,''
\newblock {\em arXiv preprint arXiv:2005.08646}, 2020.

\bibitem{nguyen2019efficient}
Van-Quang Nguyen, Masanori Suganuma, and Takayuki Okatani,
\newblock ``Efficient attention mechanism for handling all the interactions
  between many inputs with application to visual dialog,''
\newblock {\em arXiv preprint arXiv:1911.11390}, 2019.

\bibitem{tan2019lxmert}
Hao Tan and Mohit Bansal,
\newblock ``Lxmert: Learning cross-modality encoder representations from
  transformers,''
\newblock in {\em Proceedings of the 2019 Conference on Empirical Methods in
  Natural Language Processing and the 9th International Joint Conference on
  Natural Language Processing (EMNLP-IJCNLP)}, 2019, pp. 5103--5114.

\bibitem{sun2019videobert}
Chen Sun, Austin Myers, Carl Vondrick, Kevin Murphy, and Cordelia Schmid,
\newblock ``Videobert: A joint model for video and language representation
  learning,''
\newblock in {\em Proceedings of the IEEE International Conference on Computer
  Vision}, 2019, pp. 7464--7473.

\bibitem{luo2020univilm}
Huaishao Luo, Lei Ji, Botian Shi, Haoyang Huang, Nan Duan, Tianrui Li, Xilin
  Chen, and Ming Zhou,
\newblock ``Univilm: A unified video and language pre-training model for
  multimodal understanding and generation,''
\newblock {\em arXiv preprint arXiv:2002.06353}, 2020.

\bibitem{fan2019heterogeneous}
Chenyou Fan, Xiaofan Zhang, Shu Zhang, Wensheng Wang, Chi Zhang, and Heng
  Huang,
\newblock ``Heterogeneous memory enhanced multimodal attention model for video
  question answering,''
\newblock in {\em Proceedings of the IEEE Conference on Computer Vision and
  Pattern Recognition}, 2019, pp. 1999--2007.

\bibitem{li2019beyond}
Xiangpeng Li, Jingkuan Song, Lianli Gao, Xianglong Liu, Wenbing Huang, Xiangnan
  He, and Chuang Gan,
\newblock ``Beyond rnns: Positional self-attention with co-attention for video
  question answering,''
\newblock in {\em Proceedings of the AAAI Conference on Artificial
  Intelligence}, 2019, vol.~33, pp. 8658--8665.

\bibitem{liu2019roberta}
Yinhan Liu, Myle Ott, Naman Goyal, Jingfei Du, Mandar Joshi, Danqi Chen, Omer
  Levy, Mike Lewis, Luke Zettlemoyer, and Veselin Stoyanov,
\newblock ``Roberta: A robustly optimized bert pretraining approach,''
\newblock {\em arXiv preprint arXiv:1907.11692}, 2019.

\bibitem{das2017visual}
Abhishek Das, Satwik Kottur, Khushi Gupta, Avi Singh, Deshraj Yadav,
  Jos{\'e}~MF Moura, Devi Parikh, and Dhruv Batra,
\newblock ``Visual dialog,''
\newblock in {\em Proceedings of the IEEE Conference on Computer Vision and
  Pattern Recognition}, 2017, pp. 326--335.

\end{thebibliography}

\end{document}